# A transformable slender microrobot inspired by nematode parasites for interventional endovascular surgery

Xin Yang[1,2,4], Dongliang Fan[1,2,4], Yunteng Ma[1,2], Yuxuan Liao[1,2], Diancheng Li[2], U Kei Cheang[1], Bo Peng[3], Hongqiang Wang[1,2*]

[1]Department of Mechanical and Energy Engineering, Southern University of Science and Technology, Shenzhen, Guangdong, 518055, China.

[2]SUSTech Institute of Robotics Research, Southern University of Science and Technology, Shenzhen, Guangdong, 518055, China.

[3]Fuwai Hospital CAMS&PUMC, Beijing, 100037, China.

[4]These authors contributed equally to this work.

*Corresponding author: Hongqiang Wang. Email: wanghq6@sustech.edu.cn

## Abstract

Cardiovascular diseases account for around 17.9 million deaths per year globally, the treatment of which is challenging considering the confined space and complex topology of the vascular network and high risks during operations. Robots, although promising, still face the dilemma of possessing versatility or maneuverability after decades of development. Inspired by nematodes, the parasites living, feeding, and moving in the human body's vascular system, this work develops a transformable slender magnetic microrobot. Based on the experiments and analyses, we optimize the fabrication and geometry of the robot and finally create a slender prototype with an aspect ratio larger than 100 (smaller than 200 μm in diameter and longer than 20 mm in length), which possesses uniformly distributed magnetic beads on the body of an ultrathin polymer string and a big bead on the head. This prototype shows great flexibility (largest curvature 0.904 mm$^{-1}$) and locomotion capability (the maximum speed: 125 mm/s). Moreover, the nematode-inspired robot can pass through sharp turns with a radius of 0.84 mm and holes distributed in three-dimensional (3D) space. We also display the potential application in interventional surgery of the microrobot by navigating it through a narrow blood vessel mold to wrap and transport a drug (95 times heavier than the robot) by deforming the robot's slender body and releasing the drug to the aim position finally. Moreover, the robot also demonstrates the possible applications in embolization by transforming and winding itself into an aneurysms phantom and exhibits its outstanding injectability by being successfully withdrawn and injected through a medical needle (diameter: 1.2 mm) of a syringe.

**Introduction**

Cardiovascular diseases are still the leading cause of mortality worldwide among non-communicable diseases[1]. For example, the stroke incidence rate is as high as 2.5 per 1000 populations per year[2,3]. Interventional surgery is one of the most effective operations for these diseases, which utilizes the vascular system as a natural channel to approach target sites with minimum invasion. Typically, a surgeon inserts a catheter or guidewire and advances it to the target site under radiological guidance during the procedures[4]. Due to the high potentially fatal surgical risks during surgical procedures, well-training is essential for surgeons. For example, an interventional radiologist needs at least eight years of training, limiting the extensive applications of this technique and drastically raising the treatment cost[5,6]. Challenges, including navigating sharp angle turns and accessing distal small branches of blood vessels[7], still exist. Therefore, new minimally invasive or injectable surgical tools that minimize surgical risks and improve operational efficiency have gained burgeoning attention[8-12].

Recently, various robots have been proposed as alternatives, including continuum robots and small-scale untethered robots. Although continuum robots present flexibility and dexterity over their rigid-link counterparts[13], the contact area between the robot and the blood vessel wall rises as travel distance increases. The friction may confine their accessibility to the narrow and tortuous vascular networks[14,15], likely resulting in unfavored injuries to the blood vessels[5,16]. Untethered robots can move more freely without the limitation of the tether[17-20]. Among them, centimeter-scale untethered robots, typically driven by fluid pressure or functional materials (e.g., shape memory materials)[18,21], can hardly navigate freely in the complex and narrow vascular network. Although microrobots can navigate through the confined spaces[22,23], their small volume and surface area limit cargoes loading ability (cargoes usually smaller than their body sizes, e.g., a neural stem cell, 20 μm in dimension)[24,25]. Thus, the microrobot swarm has become an attractive technique since the swarm can push a heavy and large item by cooperation in groups, which is promising for drug delivery and minimum invasive surgery[26]. Still, microrobot swarms face the challenges of precisely controlling hundreds of members as desired and completely withdrawing all the robots from patients' bodies[27,28]. Being accessible to narrow and tortuous vessels, capable of carrying heavy and large cargo, and multifunctionality seem contradictory features for the design of a robot. Still, they are all essential traits for a versatile robot in treating vascular diseases.

Nature is the best inspiration source. Over 100 different parasitic worms live in human bodies[29]. They freely navigate narrow body vessels and feed, travel, and live effortlessly in the human body's blood and lymph fluid. Some intracellular and extracellular parasites can even invade the central nervous system during infection[30]. For example, as a human nematode parasite, Wuchereria bancrofti can travel in human blood vessels and lymphatic vessels, benefiting from its slender and soft body. With its small diameter, it can navigate most blood vessels, and with its long and slender body, it can transform into different shapes for different purposes[31].

Inspired by nematode parasites, we present a soft wireless magnetic robot with a microscale diameter and a centimeter-scale length (aspect ratio > 100). Due to its microscale cross-section (diameter ≤ 200 μm), it potentially can proceed inside most of the blood vessels (0.2-25 mm)[32] in human bodies (Fig. 1A). The long flexible body endows the robot with excellent adaptability and transformability for various challenging tasks (see Fig. 1B). This work designed and fabricated four candidates of nematode-inspired microrobots by the fabrication processes of thermal drawing, coating, spindle-shaped structure formation (based on Plateau-Rayleigh instability), and magnetization in sequence. Among them, the structure of beads on a string with a big magnetic head (Fig. 1, C and D) is regarded as the optimal design from experimental and theoretical comparisons. The consequent prototypes demonstrated excellent flexibility and functional variability by navigating through tortuous vessels (minimum radius of curvature: 0.84 mm), passing through narrow holes (diameter: 1.5 mm) in 3D space on demand, wrapping, delivering, and releasing a cargo that was 95 times heavier than the robot dead weight, and embolizing a simulated tumor on a blood vessel, respectively. Furthermore, we displayed that this slender robot could be successfully injected and withdrawn through a typical medical needle (diameter: 1.2 mm) of a syringe, which can benefit the interventional surgery by avoiding complications and a long healing process.

## Results

### Design and fabrication of the nematode-inspired robot

To resemble the agility of nematode parasites, we design and fabricate nematode-inspired robots in a filiform shape integrated with ferromagnetic materials, which are capable of being wirelessly driven by an external magnetic field. To optimize the performance, we conceive four structures as candidates: I) magnetic beads on a string (BOAS) with a big head, II) BOAS, III) fiber with a magnetic layer, and IV) fiber with a big head, as shown in Fig. 2A. Each of them consists of a slender fiber as the skeleton, while the ferromagnetic materials coated on the fibers are in different shapes.

To create long and ultrathin fibers for these robots, we choose the thermal drawing process, in which the diameter of the fiber $D$ is governed by[33],

$$D = \frac{C}{\sqrt{v}} \tag{1}$$

where $C$ is a constant relative to drawing conditions (e.g., the temperature and the material viscosity), and $v$ is the drawing speed. Accordingly, the filament diameter can be tuned by parameters such as temperature, drawing speed, and the viscosity of the polymer melt. Here, by increasing the drawing speed from 6 to 135 mm/s, the thermoplastic resin filament with diameters ranging from 77.61 to 622.56 μm was generated, as estimated by the above equation, as shown in Fig. 2B.

To create microstructures or uniform layers on fibers, we need to consider the dominant influences of surface tension. A magnetic liquid film on a filament spontaneously breaks into droplets[34] because the structure is unstable (Plateau-Rayleigh instability) when the axisymmetric wavelengths are more significant than the circumference of the liquid cylinder. This instability happens when the film thickness $h_t$ is larger than the critical value[35],

$$h_t = (\sqrt{2} - 1)D \tag{2}$$

The spacing λ between two adjacent magnetic beads for the BOAS structure theoretically corresponds to the diameter D of the fiber[36],

$$\lambda = (7.7 \pm 1.4)D \tag{3}$$

Thus, the spacing increases linearly as the fiber diameter increases, as shown in Fig. 2C and fig. S3 and S4. The beads' volume exhibits a similar tendency. On the other hand, whereas fiber diameter varies, the ratio of the minor axes to the major axes possesses little change, meaning that the shape of beads is relatively stable (see Fig. 2C).

According to the above analyses, we generated the prototypes of different shapes by the protocols shown in Fig. 2 D-I. Taking BOAS as an example, we first melted the thermoplastic material (hot melt adhesive 3764, 3M) at 70 °C, then dipped a needle tip into the polymer melt and drew a filament out of the hot plate (Fig. 2D). After cooling down, the fiber was immersed into a bath of a mixture (Eco-flex 0030 precursor, neodymium-iron-boron (NdFeB) microparticles, mass ratio = 1:1) until a uniform precursor layer was coated on the fiber (Fig. 2, E and F). Driven by the Plateau-Rayleigh instability effects[34], the precursor film spontaneously broke up into discrete beads, as shown in Fig. 2G. Next, this fiber was located above a permanent magnet (diameter: 100 mm, height: 10 mm) for magnetization for 6 hours at room temperature, as shown in Fig. 2H. After cured, a BOAS robot was generated, as shown in Fig. 2I. The optical images of beads formation are shown in fig. S1 and movie S1. The big magnetic head of the robots was fabricated by dipping the tip of fiber into a more viscous ferromagnetic composite precursor (Eco-flex 0030 precursor and NdFeB microparticles, mass ratio = 1:1), after which it was cured at room temperature for 50 minutes (see fig. S2). According to the dip coating theory, a more viscous precursor generates a larger bead diameter[37]. Due to the larger volume of magnetic particles and the corresponding magnetic force, the big magnetic head attached to the robots is expected to achieve better controllability.

Following similar protocols, we fabricated the fiber with a uniform composite magnetic layer using a more viscous magnetic composite precursor (the mass ratio of Eco-flex 00-30: NdFeB microparticles: silicone thinner =16:26:1) compared to the precursor for the BOAS fabrication. A more viscous liquid requires a longer duration for the liquid film to transform into beads. Therefore, enough time was available to "freeze" the uniform layer before breaking into beads. Under an external magnetic field, this layer remained on the fiber until fully cured, as shown in movie S2.

**Characterization of the nematode-inspired robots**

We compared the four prototypes of the nematode-inspired robots on two aspects: flexibility and the maximum moving speed, considering the potential applications for interventional endovascular surgeries. We first tested the bending behavior of each robot with one end fixed, as shown in Fig. 3A. Above the robot head, a permanent magnet (diameter: 10 mm, height: 10 mm, Tianming Co.) installed on a Z-axis stage was located. The characterization of the permanent magnet is shown in Section S1 and fig. S5. The initial vertical distance between the magnet and the robot was 19 mm. The magnetic field at Point O was 14.95 mT, measured by a gaussmeter (TD8620 Class 1, Tunkia Co.). Under the magnetic field, the robots bent like a cantilever. When the magnet moved down, the magnetic field on the robot became stronger, and the deflection ($\delta$) of the microrobot increased, as shown in Fig. 3B. Four prototypes of nematode-inspired robots (length: 15 mm, maximum diameter: 220 μm, the uniform beads diameter for the BOAS with a big head: 100 μm) were tested on this setup, respectively. Comparably, the deflection of the BOAS with a big head is larger than the other designs. This phenomenon can be understood by the governing equation of the deflection $\delta$ of a cantilever,

$$\delta = \frac{FL^3}{3EI} \quad (4)$$

where $F$ is the external force, $L$ is the length of the beam structure, $E$ is the elastic modulus, and I is the moment of inertia. The rigidity of BOAS with a big head is smaller than those of BOAS and fiber with a magnetic layer due to the additional volume of magnetic composite materials in the effective area and a larger cross-sectional area. On another side, the external magnetic attractive force of BOAS with a big head is stronger than that for the prototype of fiber with a big head. Thus, a stronger magnetic force and smaller bending stiffness result in the largest deflection happening on BOAS with a big head, as shown in Fig. 3B.

Moreover, we also evaluated the flexibility of the robots by measuring the bending curvature when they were in liquid and driven by a magnetic field. We put the robots in the bottom of a non-magnetic container filled with water. A permanent magnet rotated below the robots, as shown in Fig. 3C, and the robots bent accordingly. Four robots (length: 28 mm, maximum diameter: 190 μm, the diameter of the uniform beads for the BOAS with a big head: 100 μm) were employed for this test. Under different magnetic field strengths, the maximum curvatures of different robots were recorded. Compared to other robots, fiber with a big head could not bend, perhaps due to the weak driving torque. As shown in Fig. 3D, BOAS with a big head achieved the largest curvature (0.904 mm$^{-1}$ under the magnetic field of 430 mT) in this test among the four designs since it possesses a lower rigidity D and generates a smaller friction force with the container's substrate under the magnetic actuation, compared to BOAS and fiber with a magnetic layer. That means the minimum bending radius of BOAS with a big head is only 1.1 mm, and it can navigate freely through a sharp turn with such a radius. This value is approximately half of the secondary minimum bending radius (2.1 mm, BOAS). This phenomenon can be explained by the small rigidity D and adequate driving force of BOAS with a big head.

Furthermore, we analyzed and measured the movability of the robots by the maximum catching-up speed. The robot was placed in a 3D-printed non-magnetic container (polylactic acid) filled with water in the experiments, as shown in Fig. 3E. A permanent magnet (diameter: 30 mm, height: 30 mm, field strength: 27 mT, Tianming Co.) was driven by a linear slider to move forward starting from the position right below the microrobot head (Fig. 3E). Due to the guidance of the external gradient magnetic field, the robot tended to accelerate from being rest and catch up to the magnet speed. Three prototypes of BOAS with a big head (length: 27 mm) were selected for the tests. We recorded the maximum velocity that the robots could reach. In each condition, we conducted three trials. The maximum speed reached as high as 125 mm/s, as shown in Fig. 3F, which is higher than the capillary blood speed (less than 1 mm/s)[38] and the mean aortic blood speed (110 mm/s)[39]. This result means that this microrobot is promising to be employed in aortic vessels with flowing blood, one of the most challenging and dynamic environments for microrobots, in the future. Moreover, the maximum speed of BOAS with a big head is superior to other designs, as shown in fig. S6-S8, perhaps due to its advantageous trade-off between the strong driving force and weak fluidic resistance, and models can verify this tendency, which is explained in detail in Section S2. According to the above analysis and experimental results, BOAS with a big head exhibits the best flexibility and locomotion performances. Therefore, we selected this optimal design for the demonstrations.

**High adaptivity of the nematode-inspired robot in 2D and 3D environments**

To ensure successful access and efficient operation in the vascular network, the fabulous adaptivity of robots in complicated environments is highly desired. Herein, we conducted experiments in a curved channel with sharp turns to experimentally test the adaptability of our wireless nematode-inspired robot. The channel of a damped oscillation shape (the minimum radius: 0.84 mm) was infused with a mixture (volume ratio of ethanol and glycerol: 9:11) with a similar dynamic viscosity of human blood[40]. As shown in Fig. 4A and movie S3, guided by a magnetic field, the slender, nematode-inspired robot smoothly navigated through the sharp turns of this curved channel, presenting great potential for future practical applications to surgeries in tortuous and narrow blood vessels.

Moreover, the robot can also navigate in 3D space. Fig. 4B and movie S4 show three equidistant holes (diameter: 1.5 mm) of a plate mold, and the robot passed through the three holes sequentially guided by a permanent magnet, illustrating the applicability to the scenarios requiring more complex manipulation and locomotion in 3D space.

### Demonstrative application in cargo transportation in a bifurcated vascular channel

Wireless cargo transportation is critical in various medical scenarios, including minimally invasive kidney calculi removal, targeted drug delivery, and radioactive seed delivery[41,42]. Due to the microscale diameter, similar to the previous microrobots[43,44], the robot can navigate the narrow and tortuous vessels to approach the cargo. Distinct from earlier microrobots, the nematode-inspired robot in this work possesses a super large aspect ratio (over 500). The robot can transform into different configurations with its soft, slender body to grasp the targeted object or push it to the aimed site.

Here, we demonstrated that our nematode-inspired robot (length: 60 mm, body diameter: 120 μm) successfully transported the cargo, a red polymer particle of 38 mg weight (95 times heavier than the robot's body weight) and 6.7 mm in diameter (56 times larger than the robot diameter), in a bifurcated vascular channel (Fig. 5A). As shown in Fig. 5B and movie S5, the robot was guided to approach the cargo through an aorta (up to 25 mm in width) and a thin branch (2 mm in width) in sequence. Controlled by the external magnetic field, it wrapped the cargo with its long, compliant body, then carried the cargo to the aim position, and finally released the cargo for, e.g., precise treatment or removal[45].

## Demonstrative application in embolization procedure using the nematode-inspired robot

The slender nematode-inspired robot can also be applied for embolization. As a safe and effective form of therapy, embolization is widely accepted in surgeries to block the abnormal vascular vessels, including target vessel occlusion in acute hemorrhage, tumor embolization, venous varicosities, aneurysm coiling, and occluding vascular malformation[46,47]. Here, we printed an aneurysms phantom by stereolithography for tests. The aneurysm dome was 7 mm in height, and the neck diameter and parent artery diameter were both 4 mm. Actuated by an external gradient magnetic field, the robot was first guided through the blood vessel to reach the aneurysm neck and then turned its head into the dome. After several swirls, the robot ultimately entered the aneurysm and entangled inside the aneurysm (Fig. 6 and movie S6). Repeating the same procedure can induce multiple nematode-inspired robots and ultimately embolize the aneurysm to starve the cancer tissue and treat patients. Compared to previous embolization procedures by a catheter to deploy the embolizing coils through the tortuous vessels, this technique allows remote control and avoids friction and injuries on the vessels caused by the long catheter, which might result in coils inaccurate deployment[12].

## The injectability of the nematode-inspired robot

For fewer complications and a shorter healing process, medical devices' injectability is desired, which results in small incisions and few or no stitches [48,49]. Our compliant and transformable nematode-inspired robot, with a micro-size diameter and a large aspect ratio, can be carried out with a medical syringe in interventional procedures. As shown in Fig. 7 and movie S7, the prototype robot (length: 43 mm, body diameter: 75 μm) was conveniently retrieved by the medical syringe through a typical clinical needle (inner diameter: 1.2 mm) from the water during a short period (e.g., less than 8 s) only if the robot was close to the needle tip. Since the slender robot was made in one piece, unlike the microrobot swarm composed of hundreds of microrobots[24,25], it was drawn out in whole, with no debris or segment left, making this process intrinsically safe. Moreover, the flexible robot could be released from the syringe when the plunger was pushed down. During the operations of withdrawal and injection, there was no irreversible morphological damage to the robot, and the robot could maintain its function as usual (Fig. 7B).

**Conclusion**

Here, a slender nematode-inspired robot is introduced, which can travel through small-scale channels and holes and transform into various topologies for different operations. Both experiments and simulations validate that the structure of BOAS with a big head possesses a better deformation and locomotion ability than other designs. The robots exhibited their great locomotive ability by steering through a meandering channel with a sharp corner (the minimum radius: 0.84 mm) and passing through three holes in the same plane in sequence, exhibiting controllability in 3D structure. The potential applications such as cargo transportation and embolization procedures were also demonstrated on the prototypes. Furthermore, we also conducted the syringe injection and retrieval experiment, illustrating the feasibility of our approach in minimally invasive or injectable surgery.

In the future, the nematode-inspired microrobots can be improved for the more practical treatment of vascular diseases by developing an autonomous external magnetic actuation system[50], replacing the materials with biocompatible ones[51], and controlling the fabrication parameters more precisely with customized machines.

**Materials and methods**

**Fabrication of fiber with a magnetic layer:** Following the similar fabrication procedures, a magnetic layer was attached to the fiber after immersing the fiber from the composite (Eco-flex 00-30 precursor, NdFeB microparticles, silicone thinner, mass ratio, =16:26:1) and lifting it out. Then, the fiber was magnetized by a permanent magnet. Since a more viscous precursor was used for coating, a longer duration was required for beads formation. Before the surface tension broke up the magnetic composite layer, the fibers were placed onto the permanent magnet to freeze the magnetic layer shape. Therefore, a homogeneous layer of a magnetic composite layer was fabricated on the soft fiber.

**Tensile testing:** To acquire Young's modulus for the simulation, we measured the materials by a tensile testing machine (C42. 203, MTS). One of the specimens was in a rectangle shape (10 mm × 30 mm × 1 mm) composed of the ferromagnetic composite (Eco-flex 00-30 and NdFeB microparticles, mass ratio 1:1). The second specimen was in cylinder shape (diameter: 50 µm, length: 4 mm) made from thermoplastic resin (hot melt adhesive 3764, 3M). The third specimen was a fiber (diameter: 90 µm, length: 4 mm) made from the thermoplastic resin with a uniform magnetic composite layer (thickness: 20 µm). The machine stretched these three specimens at the rate of 1 mm/s to acquire their stress and strain. Finally, Young's modulus was calculated from the stress-strain curves, respectively.

**Simulation model:** Finite element models simulated the force and magnetic field in this work. We constructed the four designs of microrobots with the same maximum diameter (227 µm). The length of the robots was swept from 10 mm to 30 mm with a 5-mm step. Other conditions (such as material properties, magnetic field, and fluid characteristics) were constant. The Young's modulus of the magnetic composite and the thermoplastic resin was set as 0.4 MPa and 22 MPa, respectively.

When calculating the magnetic driving force, the boundary region was defined by a large enough sphere (radius: 40 cm) from the center of the robot, filled with air. The center of the external magnet was placed 1 cm directly in front of the head of the robots. We set the magnetization magnitude of the guiding magnet as 750 kA/m and the residual magnetic flux density of the beads as 5 mT in the Magnetic Field No Currents (mfnc) module in a steady state. The magnetic force was obtained by the sum of the forces on all the beads since the fiber was non-magnetic.

To acquire the dragging force of the robot in the fluid, the robot was fixed in a tank (100 mm × 200 mm × 100 mm), and the liquid was at a constant speed of 2 cm/s at the entrance of the channel. The no-slip condition was applied along the walls, and the pressure at the outlet was set to be zero. The gravitational force is negligible compared with drag and magnetic forces because of extremely microrobots in simulations. The fluidic resistance was computed using Laminar Flow (spf) module by integrating the total stress (including the contribution of pressure drag and viscosity resistance) over the body length in the flow direction. The density and dynamic viscosity of the fluid are 1000 kg/m$^3$ and 0.1 mPa·s, respectively.

**Magnetic actuation method in demonstrations:** For all the presented demonstrations in the main content, a cylinder NdFeB magnet (diameter: 30 mm, height: 30 mm, Tianming Co.) was used to provide the external magnetic field for actuation. The position of the magnet was placed close to the head of the robot to provide an adequate actuating force that could guide the robot. The position and posture of the robot depended on the manual movement and rotation of the magnet. For the three-dimensional manipulation experiment, the magnet was operated alternately above and below the hole so that the robot traversed the hole from top to bottom or verse visa. For other experiments, the magnet was operated under the robots.

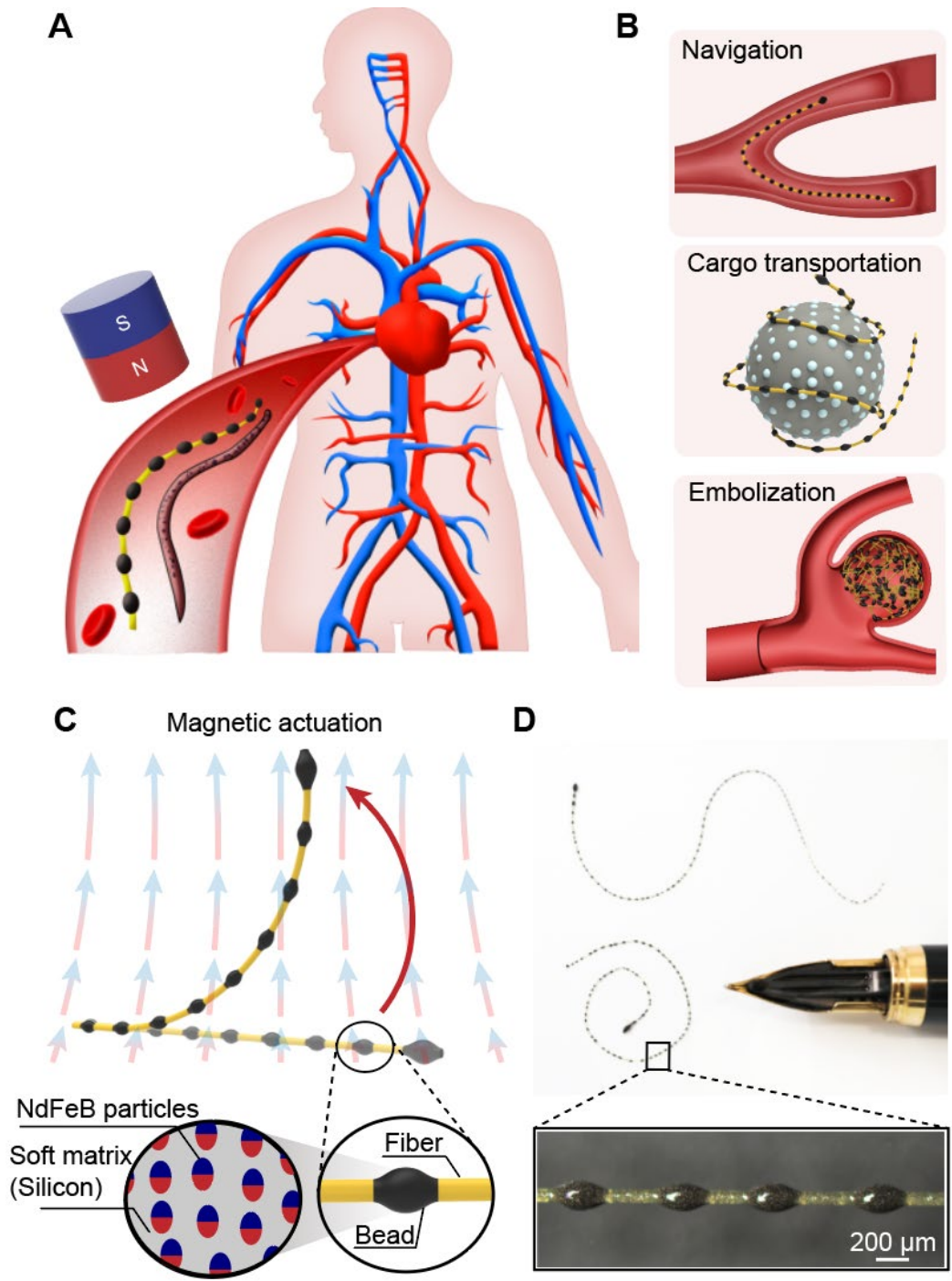

**Fig. 1. Schematic illustration of the nematode-inspired soft, slender, wireless robot**. (**A**) Imitating the structure and morphology of a nematode parasite, the slender, soft robot, actuated by external magnetic field, with multifaceted functionalities has immense potential in human vascular system. (**B**) Illustration of the potential applications of the nematode-inspired robot, including navigation, cargo delivery, and embolization. (**C**) Illustration of the active steering ability of the nematode-inspired robot resulting from the NdFeB particles embedded in the robot's magnetic beads made of silicon matrix. (**D**) The optical image of the fabricated nematode-inspired robots.

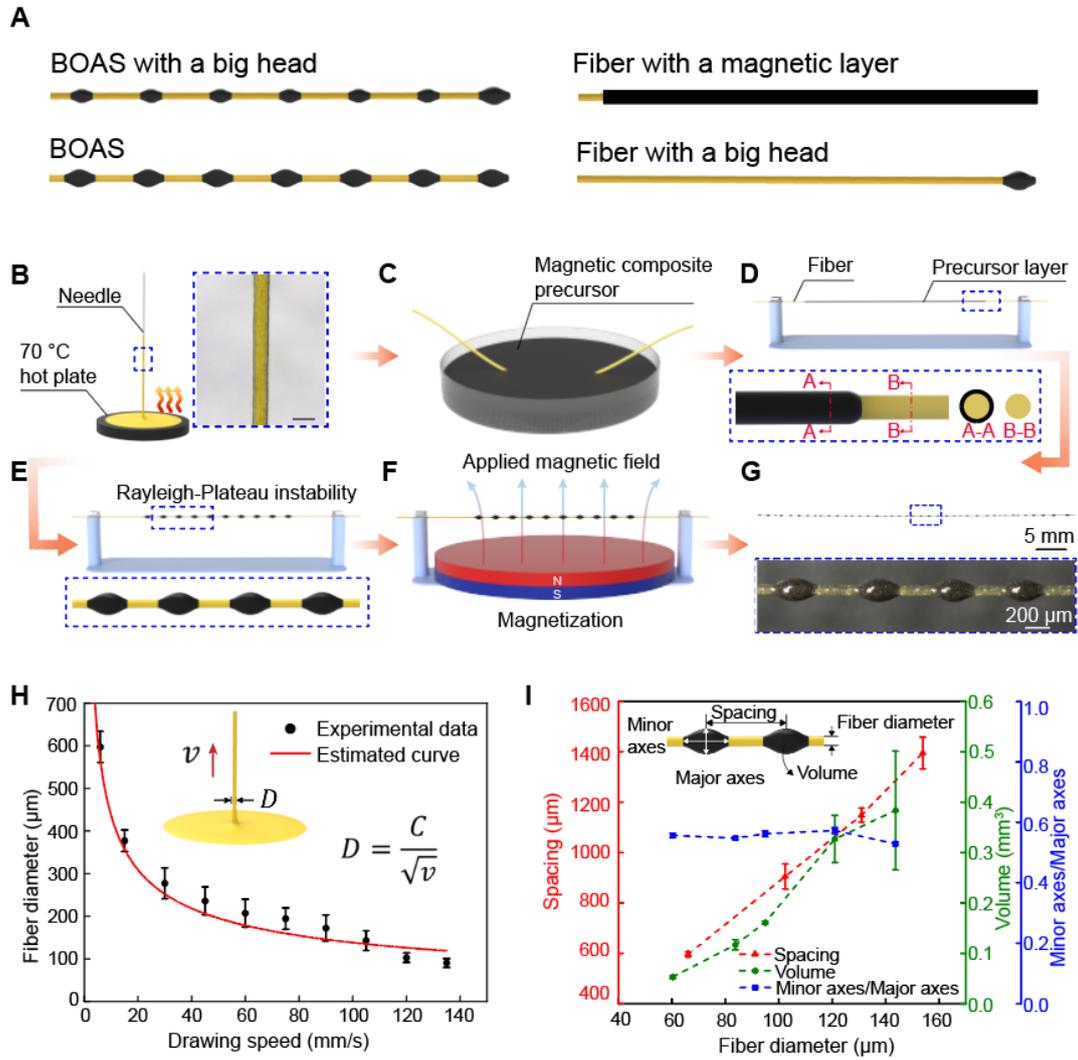

**Fig. 2. Fabrication of nematode-inspired robots.** (**A**) Schematic of the four candidate designs. (**B**) The fabrication of BOAS robot. a needle is used to draw the polymer melt from the hot plate to generate the soft filament (scale bar, 100 μm). (**C**) The filament is dipped into the magnetic composite precursor. (**D**) After extracting the filament out of the precursor, a uniform magnetic layer is coated on the filament surface. (**E**) The BOAS structure is generated in tens of seconds due to the Plateau-Rayleigh instability. (**F**) A permanent magnet is placed at the bottom of the BOAS structure for magnetization. (**G**) The magnetic BOAS structure is generated. (**H**) The relationship between drawing speed and filament diameter of the thermoplastic resin. (**I**) The morphology of the BOAS structure (spacing, magnetic bead volume and minor axes / major axes of the bead) tunning with fiber diameter.

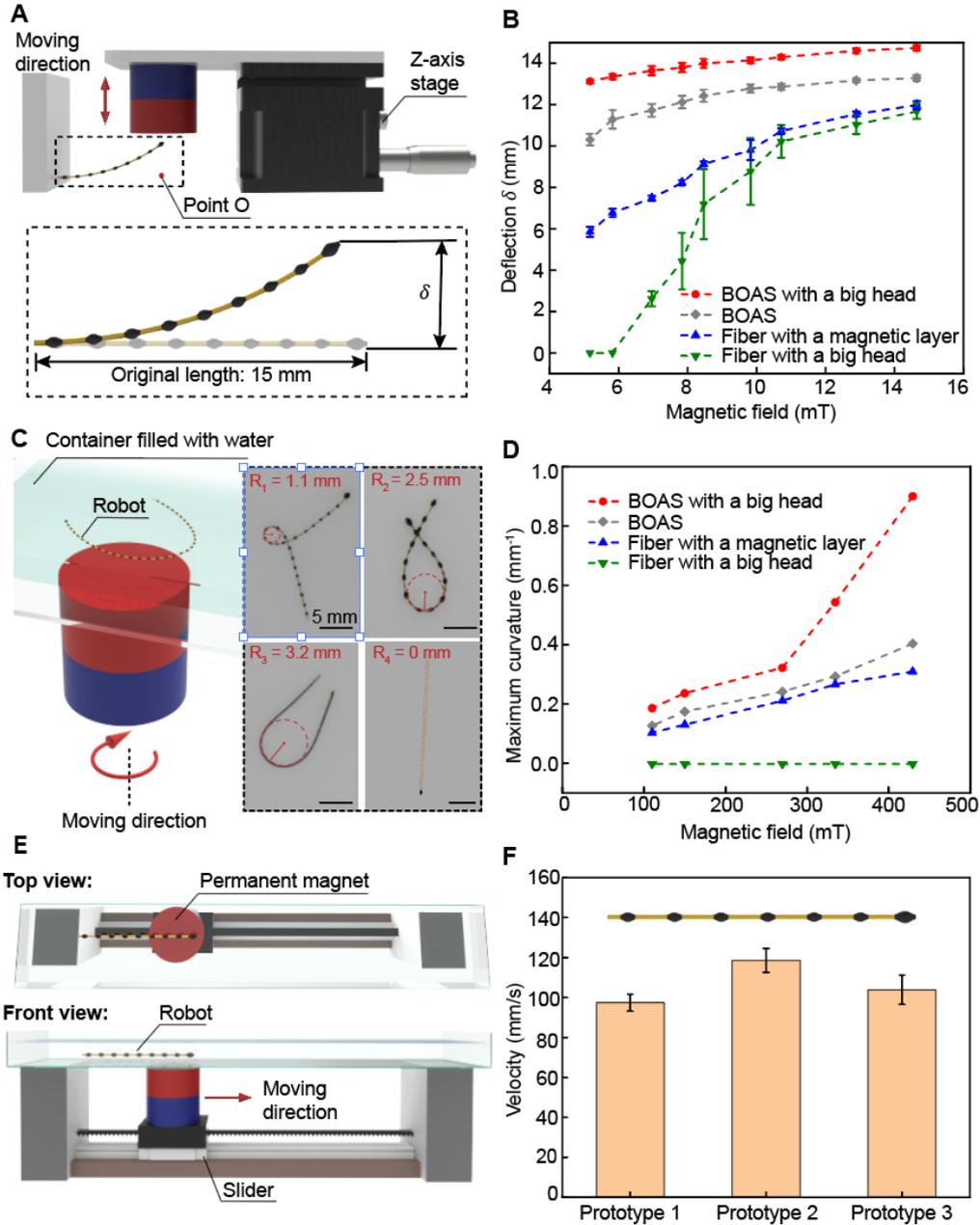

**Fig. 3 Characteristics of the nematode-inspired robots**. (**A**) Schematic of the experimental setup for deflection test. The distance between the magnet and the robot can be tuned by moving the Z-axis stage, thus changing the magnetic field. The magnetic field strength of point O is measured as the representative value of the whole field. (**B**) The bending performance of the four robot structures under different magnetic fields (measured at Point O). (**C**) Schematic of the experimental setup for curvature test (left figure) when the robot is free in the water. The minimum radius of four robot designs when the magnetic field strength is 430 mT (right figure). The permanent magnet rotates in the axial direction directly below the robot. (**D**) Curvature comparison of the four robot designs. (**E**) Schematic of the experimental setup for velocity test. (**F**) The maximum velocity of BOAS with a big head. Three prototypes are measured under the same condition.

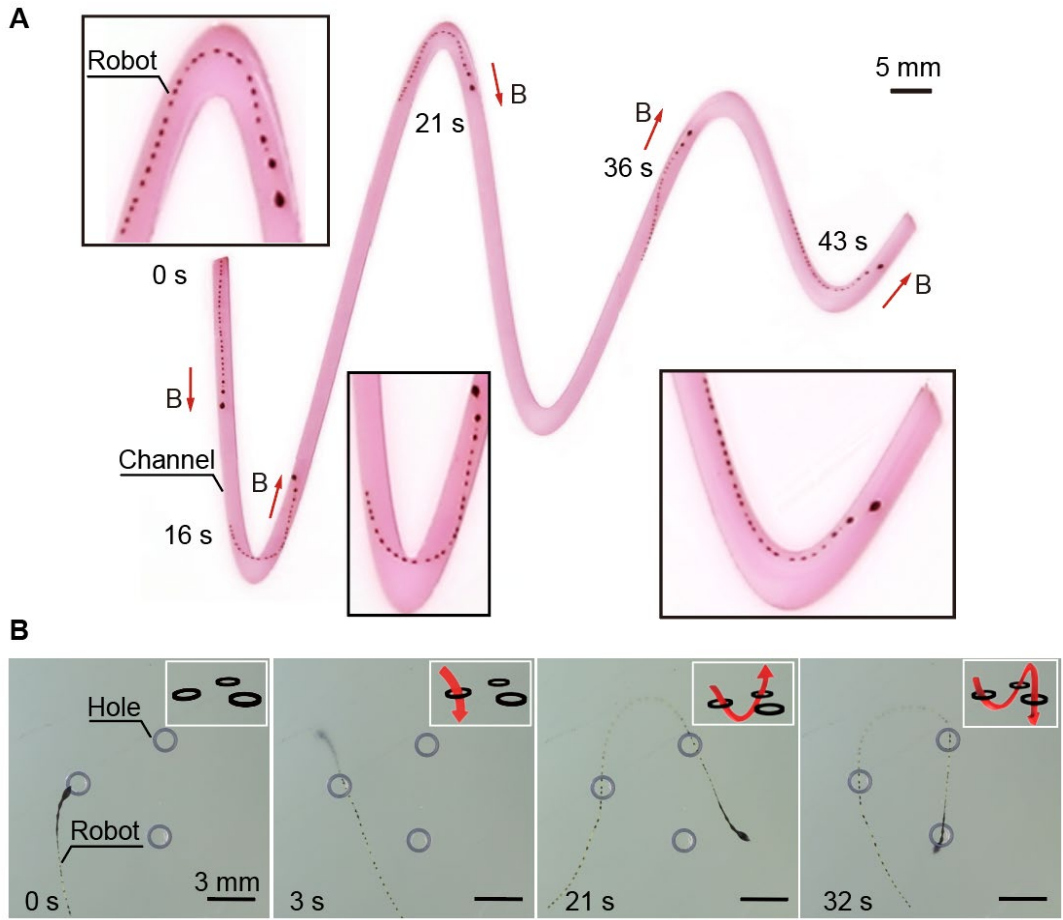

**Fig. 4 The adaptivity of the nematode-inspired robots in 2D and 3D environments**. (**A**) Experimental demonstration of navigation inside tortuous channel with high curvature. The radius of the turns from left to right are 0.84 mm, 1.02 mm, 1.19 mm, 2.20 mm, and 2.50 mm, respectively. The nematode-inspired robot smoothly navigates through this channel in 43 s. Red arrow marks the direction of applied magnetic field. (**B**) The nematode-inspired robot's locomotion in 3D space. The robot passes through the three holes sequentially from top to bottom in 32 s. The schematic representation displays the path of the magnetic head in the upper subfigure.

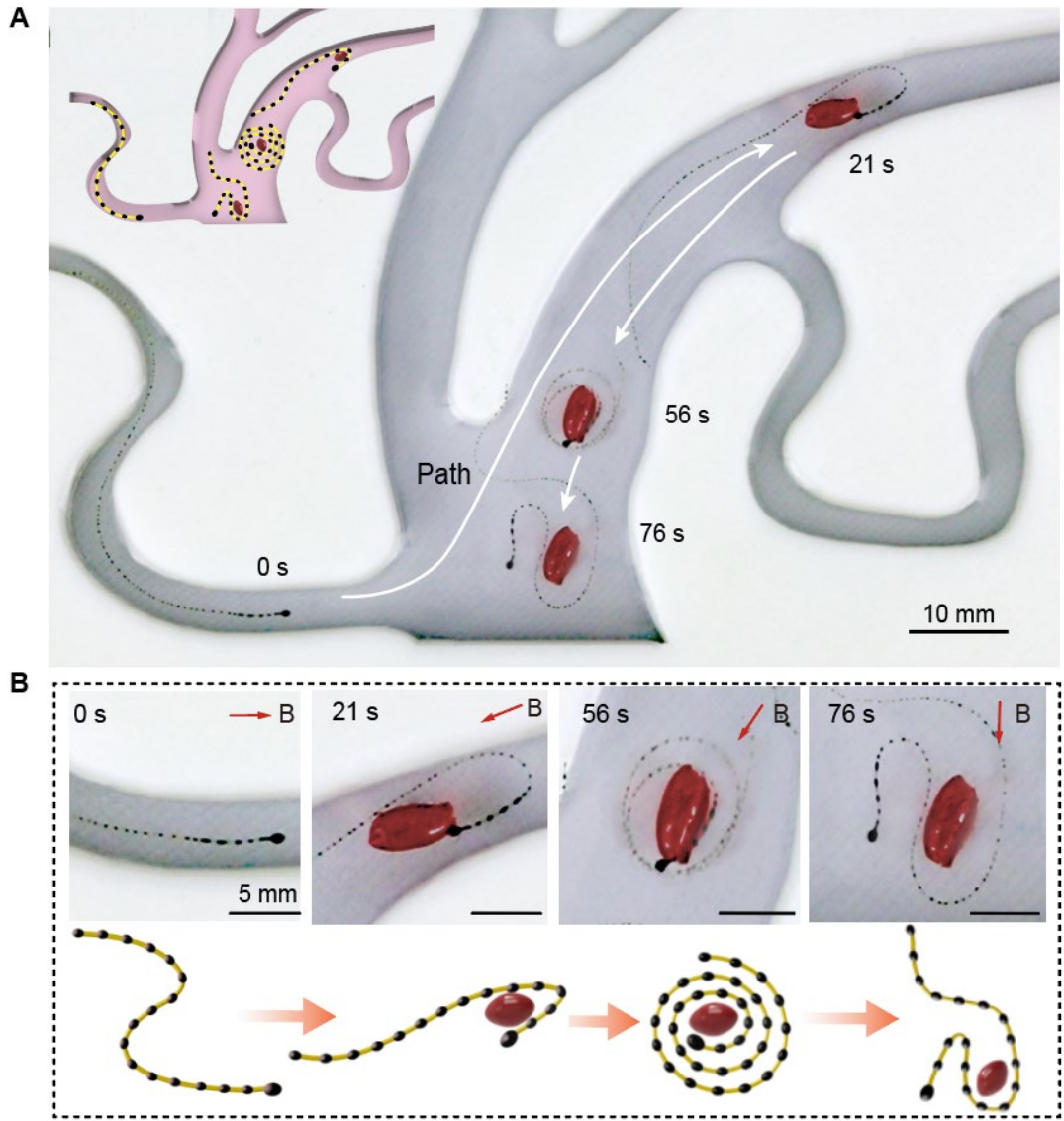

**Fig. 5 Demonstration of cargo transportation by the nematode-inspired robot in a bifurcate vascular channel.** (**A**) The overview of process in the cargo transportation process. First, the robot (57 mm long, about 120 μm in body diameter) approaches the narrow branch (2 mm in width), and then wraps the cargo (a polymer particle dyed red, 38 mg, 95 times of the body weight) by deforming itself, pushes the cargo to move, releases the cargo at the aim position, and finally returns to its starting position. White arrow marks the motion path of the robot. The upper left subfigure shows the schematic of the cargo delivery process. (**B**) Magnified images and the corresponding schematics of the cargo transportation key steps. The duration of this operation is 92 s. Red arrow marks the direction of applied magnetic filed.

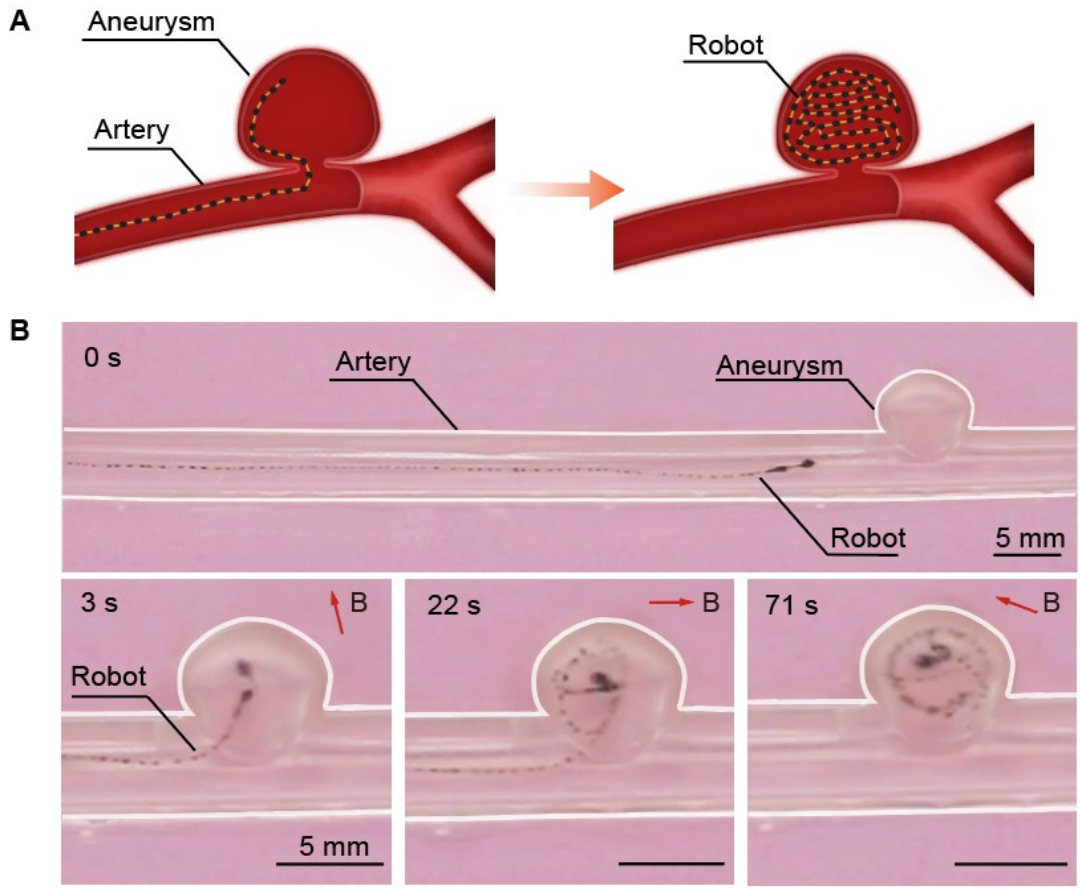

**Fig. 6 Demonstration of the nematode-inspired robot in simulated embolization**. (**A**) Schematic of the endovascular embolization procedure using the nematode-inspired robot. (**B**) Experimental demonstration of the embolization procedure. The robot first navigates through the vessel and approaches the targeted aneurysm (t = 0 s). Then the robot reaches the aneurysm neck and then turns its head into the dome (t = 3 s). By changing the direction of the applied magnetic field, the robot continues wrapping itself around (t = 22 s). After several swirls, the robot totally enters the aneurysm and entangles inside of the aneurysm (t = 71 s). The 3D aneurysm dome printed by stereolithography is 7 mm in height, with the neck diameter and parent artery diameter both 4mm. Red arrow marks the direction of the applied magnetic field.

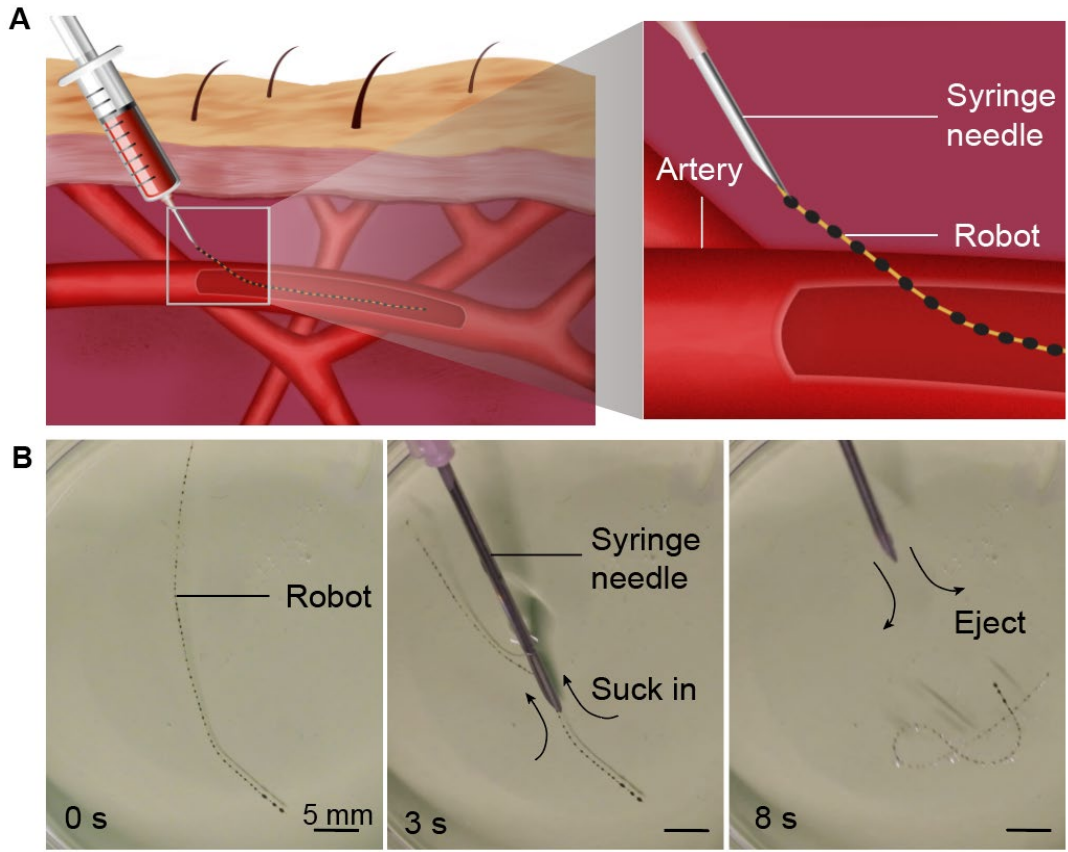

**Fig. 7 Demonstration of the injectability of the nematode-inspired robot**. (**A**) Schematic representation of the deployment and retrieval strategies of the nematode-inspired robot, delivering the robot to the targeted vessel with minimally invasive technique. (**B**) The process of syringe injection and retrieval. First, the robot (43 mm long, 75.77 μm in body diameter, 347.59 μm in head diameter) is placed in the water (t = 0 s). Under the suction of the syringe with a needle measuring 1.2 mm in inner diameter, the robot is retrieved successfully into the syringe (t = 3 s). The robot is then pushed through the needle and ejected out, with no irreversible morphological change (t = 8 s).